\documentclass[11pt,a4paper]{article}
\usepackage[hyperref]{emnlp2020}
\usepackage{times}
\usepackage{latexsym}

\usepackage{microtype}

\usepackage{graphicx}
\usepackage{amsmath}

\usepackage{algorithm}
\usepackage{algorithmicx,algpseudocode}

\usepackage{threeparttable}
\usepackage{newtxtext,newtxmath}
\usepackage{csquotes}
\usepackage{bbm}

\aclfinalcopy 

\title{BoostingBERT:Integrating Multi-Class Boosting into BERT for NLP Tasks}

\author{Tongwen Huang \\
  Tencent Corp, China \\
  towanhuang@tencent.com \\\And
  Qingyun She \\
  Sina Weibo Corp, China \\
  qingyu@staff.weibo.com \And 
  Junlin Zhang \\
  Sina Weibo Corp, China \\
  junlin6@staff.weibo.com
  \\}

\date{}

\begin{document}
\maketitle
\begin{abstract}
As a pre-trained Transformer model, BERT (Bidirectional Encoder Representations from Transformers) has achieved ground-breaking performance on multiple NLP tasks. On the other hand, Boosting is a popular ensemble learning technique which combines many base classifiers and has been demonstrated to yield better generalization performance in many machine learning tasks. Some works have indicated that ensemble of BERT can further improve the application performance. However, current ensemble approaches focus on bagging or stacking and there has not been much effort on exploring the boosting. In this work, we proposed a novel Boosting BERT model to integrate multi-class boosting into the BERT. Our proposed model uses the pre-trained Transformer as the base classifier to choose harder training sets to fine-tune and gains the benefits of both the pre-training language knowledge and boosting ensemble in NLP tasks. We evaluate the proposed model on the GLUE dataset and 3 popular Chinese NLU benchmarks. Experimental results demonstrate that our proposed model significantly outperforms BERT on all datasets and proves its effectiveness in many NLP tasks. Replacing the BERT base with RoBERTa as base classifier, BoostingBERT achieves new state-of-the-art results in several NLP Tasks. We also use knowledge distillation within the "teacher-student" framework to reduce the computational overhead and model storage of BoostingBERT while keeping its performance for practical application.
\end{abstract}

\section{Introduction}
Until recently, the NLP community is witnessing a dramatic paradigm shift toward the pre-trained deep language representation model, which achieves state of the art in various NLP tasks such as question answering \cite{CoMatchingNET,PassageRanking}, sentiment classification \cite{BERT_ASpect} and relation extraction \cite{ContextualizedWord}.

By pre-training on a large corpus on tasks such as language modeling, deep language representations are generated that can serve as a starting point to a variety of NLP tasks. Examples of pre-trained models includes ELMo \cite{peters2018deep}, ULMFiT \cite{howard2018universal}, OpenAI GPT \cite{radford2018improving} and most recently BERT \cite{devlin2018bert}. Among them, BERT outperforms its predecessors  and it consists of two stages: first, BERT is pre-trained on vast amounts of text, with an unsupervised objective of masked language modeling and next-sentence prediction. Next, this pre-trained network is then fine-tuned on task- specific labeled data.

On the other hand, ensemble is a successful approach to reduce the variance of sub-models by training multiple sub-models to combine the predictions from them. Recently, some ensemble approaches using the pre-trained transformer as the base classifier  are  proposed \cite{fajcik2019but,liutrust}. However, these works focus on bagging or stacking and there has not been much effort on exploring the boosting.

In this work, we propose a  BoostingBERT model to introduce the multi-class boosting into BERT. We all know that  some training instances with specific characteristics are more difficult to classify in many NLP tasks. We can use boosting to add extra pre-trained base Transformer and let these pre-trained base classifiers pay more attention to these difficult instances for a better task performance. In this way, BoostingBERT model  gains the benefits of both the pre-training language knowledge and  boosting ensemble.

In summary, this paper has four major contributions:
\begin{enumerate}
    \item We propose a novel model named BoostingBERT that integrates multi-class boosting into BERT. As far as we know, this work is the first one to prove that boosting can be used to enhance the performance of BERT, instead of bagging or stacking. Our experimental results also demonstrate that BoostingBERT outperforms the bagging BERT constantly.
    \item We compare two approaches making use of the base Transformer classifier in BoostingBERT model: weights privacy vs. weights sharing.   Experiment results demonstrate that the former one constantly outperforms the latter one.
    \item We conduct extensive experiments on the GLUE dataset and 3 popular Chinese NLU benchmarks.  The experiment results show that BoostingBERT significantly outperforms BERT on all tasks and demonstrates its effectiveness in many NLP tasks. We also find that BoostingBERT is  particularly useful for the tasks with little training data.
    \item Considering BoostingBERT's large number of parameters and long inference time, we use knowledge distillation within the "teacher-student" framework to reduce the computational overhead and model storage of BoostingBERT while keeping its performance.
\end{enumerate}

The rest of this paper is organized as follows. Section 2 introduces some related works which are relevant with our proposed model. We introduce our proposed Boosting BERT model in detail in Section 3. The experimental setup and results on GLUE dataset and several Chinese benchmark datasets is presented in Section 4 and Section 5. Section 6 concludes our work in this paper. 

\section{Related Work}
\subsection{Pre-trained  Models in Natural Language Processing}
Inspired from the computer vision field, where ImageNet \cite{deng2009imagenet} is used to pre-train models for other tasks \cite{huh2016makes}, many pre-trained general-purposed language encoders have generated a lot of interest in the NLP community. In contrast to fixed word embedding such as Word2Vec \cite{mikolov2013distributed} or Glove \cite{pennington2014glove}, the newer embedding generally incorporates both the left and right context.

Howard \cite{howard2018universal} proposed a general transfer learning method, Universal Language Model Fine-tuning (ULMFiT), with the key techniques for fine-tuning a language model. Peters \cite{peters2018deep} introduce embedding from Language Models (ELMo), an approach for learning high-quality, deep contextualized representations using bidirectional language models. They achieve large improvements on six different NLP tasks.

Radford \cite{radford2018improving} proposed that by generative pre-training of a language model on a diverse corpus of unlabeled text, large gains on a diverse range of tasks could be realized. BERT \cite{devlin2018bert} is the most recent inclusion to these models, where it uses a deep bidirectional transformer trained on masked language modeling and next sentence prediction objectives. It exceeds state of the art by a wide margin on multiple natural language understanding tasks.

\subsection{Ensemble Learning}
Ensemble learning uses multiple learning algorithms to obtain better predictive performance than could be obtained from any of the constituent learning algorithms alone by reducing the variance of single model. It is also one of the most popular approaches used by winners in many  machine learning competitions.
Bagging, boosting and stacking are three most widely used ensemble types.

Bagging, developed by Breiman \cite{breiman96}, is a machine-learning method that uses bootstrapping to achieve differences between models.  Models are trained on different subsets of the training data naturally through the use of resampling methods such as cross-validation and the bootstrap, designed to estimate the average performance of the model generally on unseen data. The models used in this estimation process can be combined in what is referred to as a resampling-based ensemble, such as a cross-validation ensemble or a bagging ensemble.

Boosting \cite{freund1997decision} assumes the availability of a \enquote{weak} or base learning algorithm which, given labeled training examples, produces a \enquote{weak} or base classifier. The goal of boosting is to improve the performance of the base learning algorithm. The key idea behind boosting is to choose training sets for the base classifier in such a fashion as to force it to infer something new about the data each time it is called. The fusion algorithm will finally combine many base classifiers into a single classifier whose prediction power is strong.

Stacking involves training an entirely new model to combine the predictions of several other sub-models. First, all of the  sub-models are trained using the available data, then a combiner algorithm is trained to make a final prediction using all the predictions of the  sub-models as additional inputs. 

\subsection{Ensemble Models Based on BERT}
Devlin \cite{devlin2018bert} reported that BERT shows significant increase in improvements on many NLP tasks, and subsequent studies have shown that BERT is also effective on harder tasks such as open-domain question answering \cite{CoMatchingNET}, multiple relation extraction \cite{ContextualizedWord}, and information retrieval \cite{CEDR}. 
 
 Recently, some ensemble approaches using the pre-trained transformer as the base classifier  are also proposed. However, these ensemble focus on bagging or stacking. For example,  Fajcik \cite{fajcik2019but} introduced a bagging ensemble in determining the rumor stance with pre-trained deep bidirectional transformers. The base transformer classifiers differ just by learning rate and they propose 4 different fusion approaches to increase the performance. Liu \cite{liutrust} proposed an ensemble framework to address the fake news classification challenge in ACM WSDM Cup 2019. In the proposed ensemble framework,  they trained a three-level model to perform the fake news classification. The first level of the framework contains 25 BERT model with a blending ensemble strategy. Huang\cite{huang2019} proposed a Hierarchical LSTMs for Contextual Emotion Detection (HRLCE) model to classify the emotion of an utterance given its conversational context. HRLCE is also a stacking ensemble framework which combines the results generated by BERT and HRCLE to achieve better task performance compared to BERT.

\section{Our Proposed Model}
In this section, we describe our proposed Multi-class Boosting  solution which is built on top of the pre-trained language encoder BERT in detail.

\subsection{BERT}
BERT is designed to learn deep bidirectional representations by jointly conditioning both the left and right contexts in all layers . The BERT model is pre-trained with two approaches on large-scale unlabeled text, i.e., masked language model and next sentence prediction. BERT proposes a \enquote{masked language model}(MLM) objective by masking some percentage of the input tokens at random, and predicting the masked words based on its context. The pre-trained BERT model provides a powerful context-dependent sentence representation and can be used for various target tasks, through the fine-tuning procedure, similar to how it is used for other NLP tasks.

The model architecture of BERT is a multi-layer bidirectional Transformer encoder based on the original Transformer model \cite{vaswani2017attention}. The input representation is a concatenation of WordPiece embedding \cite{wu2016google}, positional embedding and the segment embedding. 
A special classification embedding ([CLS]) is inserted as the first token and a special token ([SEP]) is added as the final token. The BERT model can also encode multiple text segments simultaneously, allowing it to make judgments about text pairs. 
To adapt BERT for specific tasks, all parameters of BERT are fine-tuned jointly by predicting a task-specific label with the task-specific output layer to maximize the log-probability of the correct label.

\subsection{Multi-Class Boosting Based on BERT}
Boosting, e.g., AdaBoost, is a popular ensemble learning technique, which combines many \enquote{weak} or base classifiers and has demonstrated to yield better generalization performance in many applications. Boosting takes as input a training set $\mathcal{D}=\{(x_1,y_1),\cdots,(x_i,y_i),\cdots, (x_m,y_m)$\}, where each $x_i$ belongs to some instance space X, and each label $y_i$ is in some label set Y. AdaBoost calls a given \enquote{weak} or base learning algorithm repeatedly in a series of rounds $t=1, \cdots, i , \cdots, T $. One of the main ideas of the boosting is to maintain a distribution or set of weights over the training set. The weight of this distribution on training example $i$ on round $t$ is denoted $\mathcal{D}_t(i)$. Initially, all weights are set equally, but on each round, the weights of incorrectly classified examples are increased so that the \enquote{weak} or base learner is forced to focus on the hard examples in the training set.

We intent to combine the Boosting with BERT in this work. Though original AdaBoost is a successful technique for solving the two-class classification problem, most NLP tasks belong to multi-class classification problem. So a multi-class version boosting algorithm is needed firstly for further work. On the other hand, the classic base classifier fusion strategy of AdaBoost is the weighted voting by base classifiers. We deem it's relatively simple to effectively ensemble the complex base models such as BERT and we need a more complicated fusion approach. So we design BoostingBERT as a two stages process:  The first stage introduces the multi-class classification ability into BERT to train the pre-trained Transformer  classifiers in boosting way; In the second stage, we train a fusion network to ensemble each Transformer encoder after fixing the parameters of base classifiers. Because the performance of fusion network is comparable with or slightly better than the commonly used weighted voting strategy, so we use the strategy during this work.
\subsubsection{First Stage: Base Classifier Training}
In going from two-class to multi-class classification, most boosting algorithms have been restricted to reducing the multi-class classification problem to multiple two-class problems. Zhu \cite{zhu2009} proposed a new algorithm that directly extends the AdaBoost algorithm to the multi-class case without reducing it to multiple two-class problems. Surprisingly, this algorithm is almost identical to AdaBoost but with a simple modification. Inspired by Zhu's work, we follow similar idea and improve this multi-class boosting algorithm further. BoostingBERT introduces  BERT into boosting framework in weight initialization phrase as algorithm 1 shows.

Algorithm \ref{algorithm:a1} shares the similar simple modular structure of AdaBoost with several differences. As for going from two-class to multi-class classification, algorithm 1  adds the extra term $log(K - 1)$ to resolve this problem just as Zhu's work does. While K is the label number. In order to introduce the pre-trained language encoder BERT into the boosting, we  initialize the parameters of  Transformer  according to some kind of weight initialization approach when adding a new base classifier into ensemble. Various weight initialization strategies can be adopted here and we will explain them in detail in the  appendix page because of the limited space.

Another improvement for the multi-class boosting BERT is to multiply the new weight $w_i$ to each instance's loss without re-normalizing it in training process in step 2.6. In this way, BoostingBERT pays more attention to hard examples by adding a new pre-trained base Transformer. We have tested the re-normalizing approach and found the performance decreases with a large margin.

\begin{algorithm}[H]
\caption{Multi-Class Boosting BERT}
\label{algorithm:a1}
1. Initialize the training instance weights $w_i=1/n, i=1,2,\cdots, n$.
\\
2. For m=1 to M:

(2.1) Initialize the weights of  Transformer encoder $T^{(m)}(x)$ according to specific weight initialization strategy.

(2.2) Fine-tuning a Transformer encoder $T^{(m)}(x)$ to the training data using weights $\{w_i\}_{i=1}^{n}$.

(2.3) Compute
$$
err^{ ( m ) } = \sum _ { i = 1 } ^ { n } w _ { i } \mathbbm{1} \left( c _ { i } \neq T ^ { ( m ) } \left( \boldsymbol { x } _ { i } \right) \right) / \sum _ { i = 1 } ^ { n } w _ { i }
$$
(2.4) Compute
$$
\alpha ^ { ( m ) } = \log \frac { 1 - err ^ { ( m ) } } {err^{(m)}} + \log(K-1)
$$

(2.5) Set
$$
w_{ i } \leftarrow w _ { i } \cdot \exp \left( \alpha ^ { ( m ) } \cdot \mathbbm{1} \left( c _ { i } \neq T ^ { ( m ) } \left( \boldsymbol { x } _ { i } \right) \right) \right)
$$

for $i=1,\cdots,n$.

(2.6) Multiply $w_i$ to the loss of instance $x_i$
\\ 
3. Fusion network training.\\
(3.1) Fix the parameters of each base classifier;
\\ 
(3.2) For each training  instance $x_i$ in training data.
\\ 
(3.2.1) For m=1 to M :

\setlength{\parindent}{2em}(3.2.1.1) Pass the $x_i$ to the base classifier Transformer $T^{(m)}$ and output the softmax distribution $p_{soft}^{m}$.
$$
p_{soft}^{m} = T^{(m)}(x_i)
$$

\setlength{\parindent}{2em}(3.2.1.2) Multiply $\alpha ^ { ( m ) }$ to the softmax distribution  $p_{soft}^{m}$ of the $m$-th base classifier $T^{(m)}$.
$$
p_{soft}^{m} = \alpha ^ { ( m ) } \cdot T^{(m)}(x_i)
$$

\setlength{\parindent}{2em}(3.2.1.3) Concatenate the current softmax distribution
$$
p_{soft}=Concate(p_{soft},p_{soft}^{m})
$$

\noindent(3.2.2) Train the parameters of MLP layers on top of $p_{soft}$.
\\ 
4. Output: Predict class label by fusion network.
\end{algorithm}

\subsubsection{Second Stage:Fusion Network Training}
In order to better ensemble each strong BERT base classifier, the fusion network is adopted on the top of each Transformer encoder. The parameters of each base classifier is fixed during the training of the fusion network. The fusion network consists of several hidden layers of MLP network which use the output of each Transformer encoder as input layer. Because the base classifier has various confidence score on specific task, we multiply this weight $\alpha$ of each base classifier to its output to use this information. The specific training procedures of fusion network are described in step 3 of the multi-class BoostingBERT algorithm.

\subsection{Weights Sharing BoostingBERT}

\begin{figure}[hbt!]
\includegraphics[width=7.5cm, height=6cm]{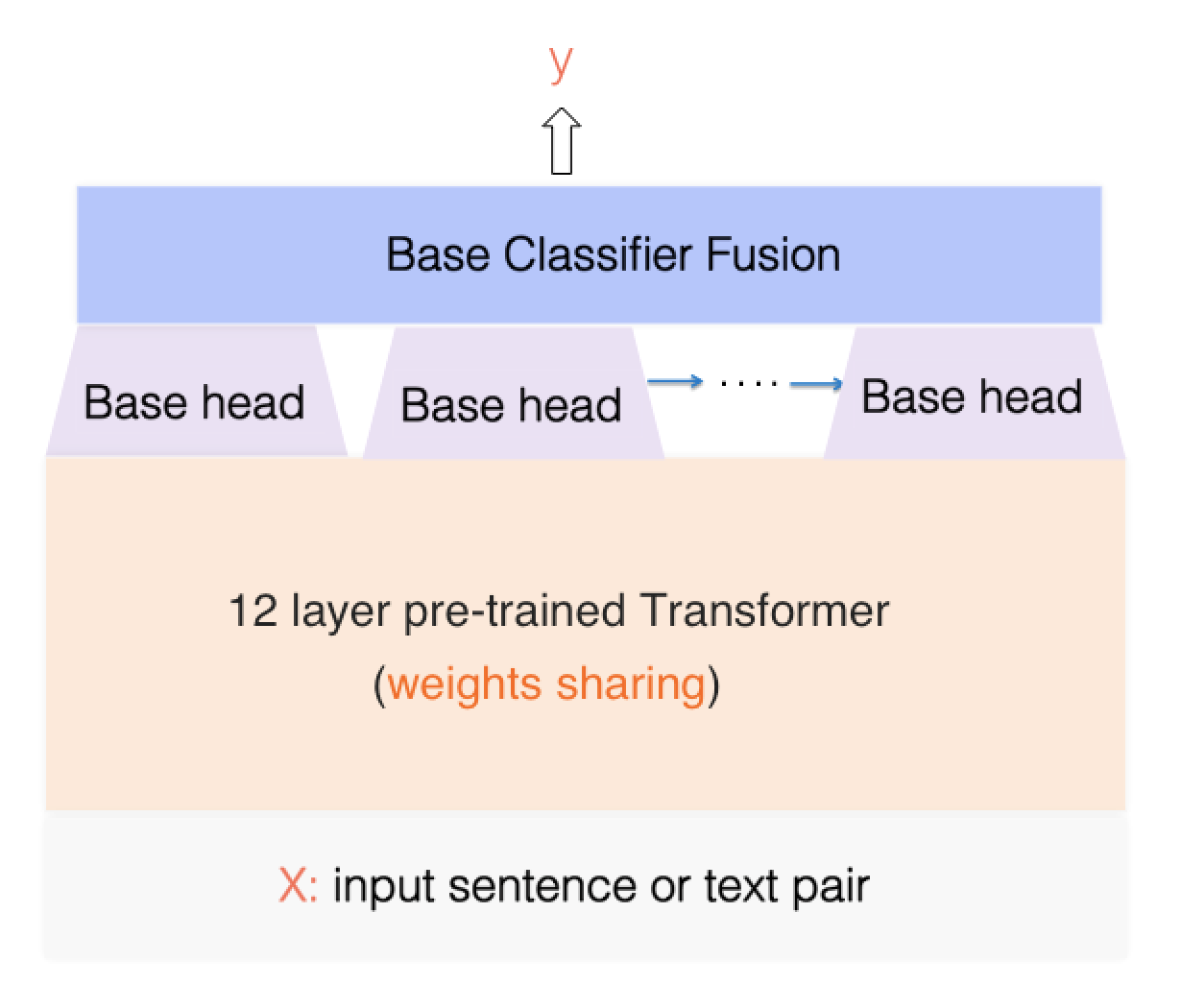}
\caption{Weights Sharing BoostingBERT}
\label{fig:f1}
\end{figure}

\begin{figure*}[h]
\centering
\includegraphics[width=12cm,height=6cm]{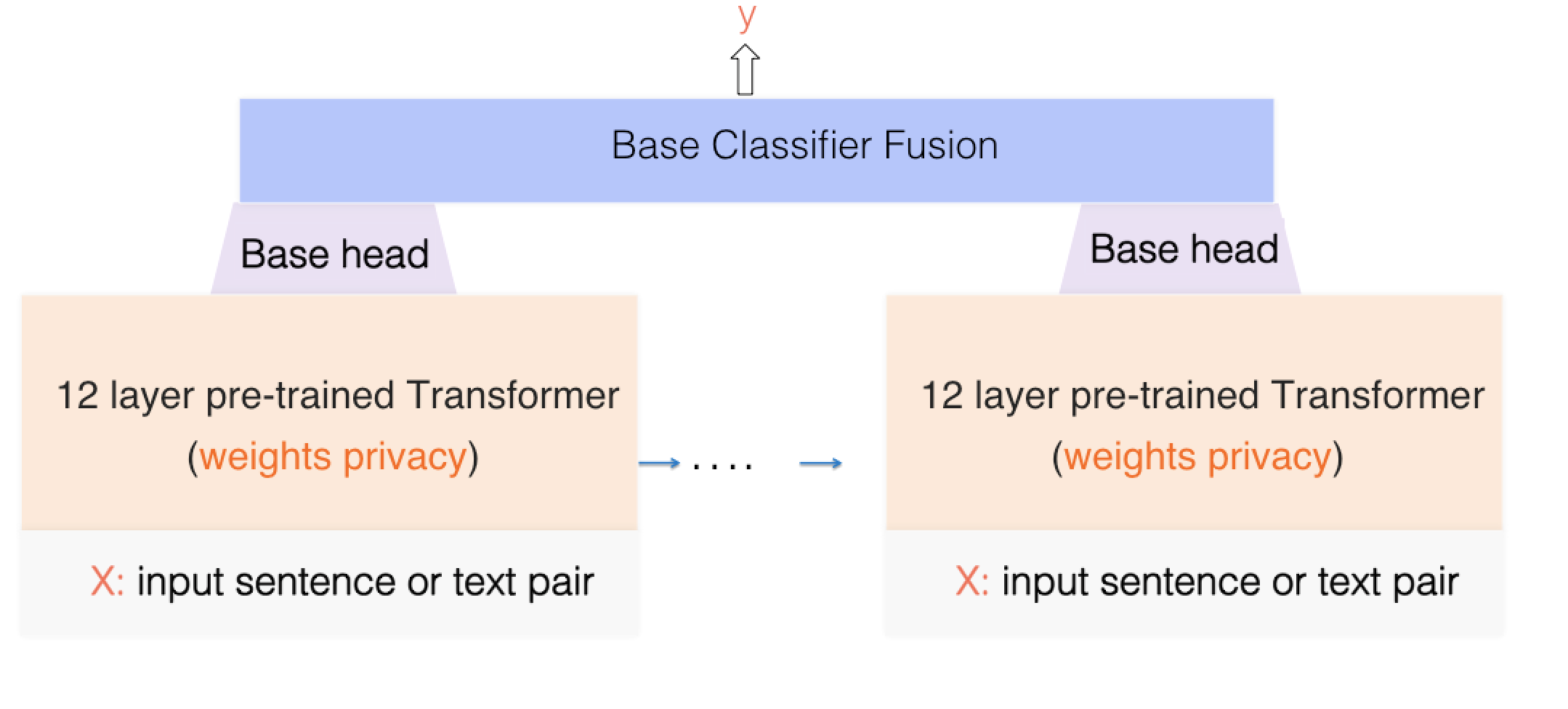}

\caption{Weights Privacy BoostingBERT}
\label{fig:f2}
\end{figure*}

The base classifier in BoostingBERT is a 12 layer pre-trained Transformer. When a new Transformer is added into boosting, the weights of  Transformer  $T^{(m)}(x)$ is initialized according to specific weight initialization strategy. Then BoostingBERT model chooses hard training instances  to fine-tune the new Transformer to force it to encode new features for these hard examples.

Consisting of many base classifiers to fine-tune,  weights sharing BoostingBERT (Figure \ref{fig:f1}) considers learning a general-purpose model that shares most parameters across base Transformers just like multi-task learning approach does. The linear head which has a small number of specific parameters can be added to adapt the shared model for each base classifier. On the contrary, the weights privacy BoostingBERT does not share any parameter across base classifiers(Figure \ref{fig:f2}). Obviously, weights sharing across base classifiers need much fewer parameters compared to weights privacy strategy and it opts to avoid overfitting. Surprisingly, our experimental results show that weights privacy outperforms weights sharing strategy constantly. We will compare two model's performance difference in detail in section 5.

\subsection{Knowledge Distillation of BoostingBERT}
Consisting of many base classifiers, BoostingBERT has an extremely large number of parameters and needs a long time to inference. This means that it's difficult to be deployed in real life applications. In order to make BoostingBERT faster and smaller, we use knowledge distillation to compress it.

Knowledge distillation is a model compression technique in which a compact "student" model is trained to reproduce the behaviour of a larger "teacher" model or an ensemble of models. It improves training because the full distribution over labels provided by the teacher provides a richer training signal than a one-hot label. So the student can accelerate deep model inference and reduce model size while maintaining accuracy.

In this work, the student has the same general architecture as a single BERTBase model while the BoostingBERT model is used as the teacher. Knowledge distillation trains the student to instead match the predictions of a teacher model with the following loss:
\begin{equation}
\mathcal{L}_{s}(\theta) = \sum_{(x_i,y_i) \in \mathcal{D}}l(f_{t}(x_i), f_{s}(x_i, \theta))    
\end{equation}
where $f_s$ and $f_t$ denote the mapping functions of student and teacher model.

To further improve the student model's ability, we mixes the teacher prediction with the gold label during training just as "teacher annealing"\cite{clark2019bam} does. 
\begin{equation}
\begin{aligned}
\mathcal{L}_{s}(\theta) = \sum_{(x_i,y_i) \in \mathcal{D}} &  \bigg ( \lambda \cdot l(y_i, f_{s}(x_i, \theta)) \\ 
&+(1-\lambda) \cdot l(f_{t}(x_i), f_{s}(x_i, \theta)) \bigg )
\end{aligned}
\end{equation}
where $\lambda$ is linearly increased from 0 to 1 throughout training. In the early training stage, the model is mostly distilling to get as useful of a training signal as possible from teacher. Towards the end of training, the model is mostly relying on the gold-standard labels so it can learn to try to surpass its teachers.

\section{Experimental Setup}

\subsection{Datasets}
We evaluate the proposed BoostingBERT model on the GLUE dataset and 3 popular Chinese NLU benchmarks: ChnSenti, Cross-lingual Natural Language Inference corpus (XNLI), and ECDT dataset. More datasets details are described in appendix page because of the limited space.

\subsection{Implementation details}
Our implementation of Boosting BERT is based on the TensorFlow and the BERT base is used as base classifier of ensemble . We use the uncased English BERTbase model for English tasks and BERTchinese for Chinese tasks. The Adam is used as our optimizer with a learning rate of 2e-5 and a batch size of 32. we set maximum number of epochs to 3 with a dropout rate of 0.1 and the maximum number of base classifiers to 9.

\section{Experimental Results}
We use the BERT base fine-tuning model as one strong baseline, which consists of 12-layer transformer blocks, 768 hidden size, and 12 heads. Several BERT base fine-tuning models are trained using different hyper-parameters. We compare  our proposed multi-class boosting BERT model with this strong baseline and demonstrate the effectiveness of  the boosting  BERT.

\subsection{GLUE Results}
\begin{table*}[hbt!]
\label{table:t3}
\caption{The overall results of GLUE dataset on dev set}
\centering
\begin{tabular}{lllllllll}
Model        & CoLA  & SST-2 & MRPC  & QQP   & MNLI$_{m}$ & MNLI$_{mm}$ & QNLI  & RTE   \\ \hline
BERT base    & 80.72 & 92.55 & 85.29 & 90.97 & 84.39  & 85.02   & 91.56 & 67.15 \\
BoostingBERT & 82.93 & 93.35 & 87.87 & 91.59 & 85.16  & 85.64   & 92.04 & 72.92 \\ \hline
$\Delta$ & +2.21  & +0.8   & +2.58  & +0.62  & +0.77  & +0.62   & +0.48 & +5.77 \\ \hline
\end{tabular}
\end{table*}

Table 1 gives the overall results on GLUE. The first observation is that our model architecture achieves better results compared to BERT. We can see that multi-class boosting BERT outperforms BERT on all tasks, which means the boosting helps correctly classifying the difficult instances by adding extra base BERT classifier in many types of NLP tasks.

The experimental results also show that boosting BERT is particularly useful for the tasks with little training data. As we observe in the Table 1, the improvements over BERT are much more substantial for the tasks with less training data e.g., RTE, MRPC and COLA.

\begin{figure}[hbt!]
\includegraphics[width=8cm, height=6cm]{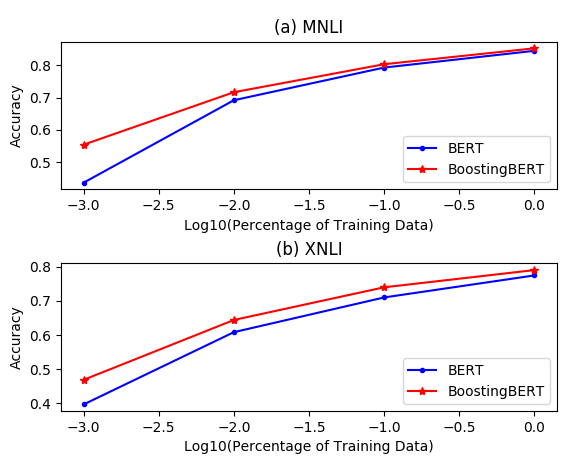}
\caption{Results on MNLI and XNLI with different ratios of training data. The X-axis indicates the amount of samples used for training.}
\label{fig:f3}
\end{figure}

\begin{table}[!htbp]
\centering
\caption{Results on MNLI and XNLI with different ratios of training data, as shown in Figure~\ref{fig:f3}. The model name ``B2BERT'' means BoostingBERT.}
\label{table:t40}
\begin{tabular}{lllll}
Model & 0.10\% & 1\% & 10\% & 100\% \\\hline
\multicolumn{5}{c}{MNLI Dataset (Dev Accuracy)} \\ \hline
\# Data & 392 & 3.9K & 39K & 392K \\ \hline
BERT & 43.75 & 69.16 & 79.23 & 84.39 \\
B2BERT & 55.46 & 71.63 & 80.26 & 85.16 \\\hline
$\Delta$ & +26.77 & +3.57 & +1.30 & +0.91 \\ \hline
\multicolumn{5}{c}{XNLI Dataset (Dev Accuracy)} \\ \hline
\# Data & 392 & 3.9K & 39K & 392K \\\hline
BERT & 39.60 & 60.84 & 71.04 & 77.51 \\
B2BERT & 46.83 & 64.42 & 74.02 & 79.08 \\ \hline
$\Delta$ & +18.26 & +5.88 & +4.19 & +2.03 \\ \hline
\end{tabular}
\end{table}

To verify the assumption that BoostingBERT performs better on less training data tasks, we conduct extra experiments on two tasks, MNLI and XNLI. The training data of each task is split and randomly sampled 0.1\%,1\%,10\% and 100\% as four  new training data. The corresponding experimental results on different amounts of training data of MNLI and XNLI are shown in Figure \ref{fig:f3} and Table \ref{table:t40}. We can see from the results that the fewer training data indeed brings larger performance improvement in both tasks. For example, with only 0.1\% of the MNLI training data, BoostingBERT achieves 26.77\% performance gains compared with BERT; while with 10\% of training data, the performance gain decreases to 1.3\%. The similar results can also be observed on XNLI task.

We deem that the main reason lies in the fact that the harder instances in training data provide much more information for the task. So the boosting's advantage in paying more attention to difficult instances increase the generalization ability of model, especially for small training data tasks.

\subsection{Chinese  Datasets Results}

\begin{table}[!htbp]
\caption{Chinese dataset dev set results. The model name ``B2BERT'' means BoostingBERT}
\label{table:t4}
\begin{center}
\begin{tabular}{lllll}
\hline
Model        & ECDT & ChnSenti & XNLI & Score \\ \hline
BERTBase     & 94.54& 93.00 & 77.51 & 88.35 \\ 
B2BERT & 96.88& 94.33 & 79.08 & 90.1  \\ \hline
$\Delta$        & +2.34 & +1.33  & +1.57  & +1.75  \\ \hline
\end{tabular}
\end{center}
\end{table}

The test results on 3 Chinese NLP tasks are presented in Table \ref{table:t4}. It can be seen that multi-class boosting BERT outperforms BERT on all tasks significantly. For all NLP tasks, we obtain more than 1\% absolute accuracy improvement over BERT.  Among all three tasks, two of them are also tasks with little training data. We attribute the performance gain to the boosting's ability of paying more attention to the difficult training instances.

\subsection{The Impact of Parameter Sharing}

\begin{table}[!htbp]
\caption{GLUE tasks dev set results(WS means weight sharing and Non-WS means weight privacy approach)}
\label{table:t6}
\begin{tabular}{lllll}
\hline
Model & MRPC & MNLI$_{m}$ & RTE & Score \\ \hline
WS & 86.27 & 84.54 & 71.12 & 80.64 \\
Non-WS & 87.87 & 85.16 & 72.92 & 81.98 \\ \hline
\end{tabular}
\end{table}

\begin{table}[!htbp]
\caption{Chinese dataset dev set results(WS means weight sharing and Non-WS means weight privacy approach)}
\label{table:t7}
\begin{tabular}{lllll}
\hline
Model & ECDT & ChnSenti & XNLI & Score \\ \hline
WS & 96.49 & 94.33 & 77.83 & 89.55 \\
Non-WS & 96.88 & 94.33 & 79.08 & 90.10 \\ \hline
\end{tabular}
\end{table}

We design some experiments on three typical GLUE tasks and three Chinese datasets to verify the effectiveness of weight sharing of base Transformer classifier in multi-class boosting BERT model. Results are presented in Table \ref{table:t6} and Table \ref{table:t7}. We can see that weight privacy strategy constantly outperforms weight sharing in almost all NLP tasks, no matter the English corpus or Chinese corpus.

The experimental results seems surprising because the weight sharing approach obviously has much fewer parameters to learn and is opt to avoid overfitting problem compared to weight privacy strategy. Jawahar\cite{jawahar2019does} shows that the BERT requires deeper layers for handling harder cases involving long-distance dependency information. According to their research conclusion, we suppose the reason behind this surprising result lies in that the base classifiers in BoostingBERT focus on the difficult instance in training data and the deeper layers of  Transformer encode those features. So the deep layers of Transformer in weight privacy strategy keep the features and weight sharing of different Transformers weakens this ability.

\subsection{The Effect of Knowledge Distillation}

\begin{table}[!htbp]
\caption{Performance comparison of BoostingBERT teacher(B2BERT) and Knowledge Distillation student(B2B-KD).}
\label{table:10}
\begin{tabular}{lllll}
Model & ChnSenti & ECDT & SST2 & RTE  \\\hline
BERTBase & 93.00 & 94.54 & 92.55 & 67.15  \\
B2BERT & 94.33 & 96.88 & 93.35 & 72.92  \\
B2B-KD & 94.67 & 96.49 & 93.00 & 72.56 \\ \hline
\end{tabular}
\end{table}

We conduct experiments on two Chinese dataset and two English datasets to testify the effect of knowledge distillation of BoostingBERT. From the experimental results showed in Table \ref{table:10}, we see that: 1) KD student is consistently better than BERTBase in all tasks and achieves improvement of 3.1\% on average. 2) Compared with the teacher BoostingBERT, KD student maintains comparable performances while the model is much smaller and faster. These observations indicate that BoostingBERT can be applied in real life applications through knowledge distillation.

\subsection{Comparison with Bagging BERT}
\begin{table}[!htbp]
\caption{Performance comparison of Bagging BERT and BoostingBERT.}
\label{table:11}
\begin{tabular}{llll}
Model & BERTBase & BagBERT & B2BERT \\ \hline
MRPC & 85.29 & 86.52 & 87.87 \\
MNLI$_{m}$ & 84.39 & 84.61 & 85.16 \\
RTE & 67.15 & 71.12 & 72.92 \\ \hline
ECDT & 94.54 & 96.10 & 96.88 \\
ChnSent & 93.00 & 93.5 & 94.33 \\
XNLI & 77.51 & 77.63 & 79.08  \\ \hline
\end{tabular}
\end{table}
Bagging is another commonly used ensemble which also considers homogeneous weak learners just as boosting does. Bagging often learns base learners independently from each other in parallel and combines them following some kind of deterministic averaging process while boosting learns them sequentially in adaptative way.
We design some experiments on six datasets to compare the performance of two different ensembles. As for the bagging BERT, the base BERT classifiers are trained using different learning rate. We combine the predictions of all base models above by taking the unweighted average of the posterior probabilities for these models and the final prediction is the class with the largest averaged probability.
Experimental results are presented in Table \ref{table:11} and we can see from these results that BoostingBERT outperforms Bagging BERT constantly.

\subsection{Replacing the base learner with RoBERTa}

\begin{table}[!htbp]
\caption{Performance comparison of BoostingBERT with RoBERTa base classifier(BRoBERTa) and Best Model published.}
\label{table:12}
\begin{threeparttable}
\begin{tabular}{lllll}
Model & MNLI$_{m}$ &MNLI$_{mm}$ &  SST2  \\\hline
BestSingle\tnote{*} & 87.23 & 87.12 & 93.81 \\\hline
BestEnsemble\tnote{*} & 86.22 & 86.53 & 93.46 \\\hline
RoBERTaBase\tnote{**} & 84.3 & - &  92.90  \\ \hline
RoBERTaBase & 86.68 & 86.69 & 93.12  \\ 
BRoBERTa & 87.47 & 87.31 & 93.87  \\ \hline

\end{tabular}
 \begin{tablenotes}
        \footnotesize
        \item[*] {Results from Google T5 \cite{raffel2019exploring}.} 
        \item[**] {Results from Table 2 in RoBERTa \cite{liu2019roberta}} 
      \end{tablenotes}
\end{threeparttable}
\end{table}

Since the emergence of BERT, many new stronger models have been proposed recently. We will show that BootingBERT can significantly  boost the performance of NLP tasks if we replace the base learner with stronger base classifier. RoBERTa\cite{liu2019roberta} shows that training BERT longer on more data leads to significant boost in performance. We use the RoBERTa with 12 layers transformer blocks as BoostingBERT's base classifier instead of BERTBase and design some experiments. Table \ref{table:12} shows the experimental results. We also compare results of BoostingBERT with the best model performance reported in public papers in several NLP tasks , which came from google T5\cite{raffel2019exploring} model(both the best single model and ensemble model among 70 different models with various configurations). T5 model compares various pre-training objectives, architectures, transfer approaches, and other factors on dozens of language understanding tasks and introduces a much larger new "Colossal Clean Crawled Corpus" in pre-training process. We can see from the results that BoostingBERT achieves new state-of-the-art results in several NLP tasks.

\section{Conclusion}
In this paper, we propose the  BoostingBERT model which is a new approach integrating multi-class boosting into BERT. We perform an extensive number of experiments on the GLUE dataset and three Chinese NLU benchmarks.  The experiment results show that BoostingBERT outperforms strong BERT  baseline on all tasks and demonstrates its effectiveness in various NLP tasks. We also find that BoostingBERT is  particularly useful for the tasks with little training data. Two approaches making use of the base Transformer classifier in BoostingBERT model are also compared and experiment results demonstrate that the weight privacy strategy constantly outperforms the weight sharing approach.


\bibliographystyle{acl_natbib}
\bibliography{anthology,emnlp2020}
\appendix

\newpage
\section{Supplemental Material}
\subsection{Dataset description}
\subsubsection{GLUE Benchmark}
\begin{table}[!htbp]
\caption{Summary of GLUE benchmark}
\label{table:t1}
\begin{tabular}{llllc}
Corpus & \#Train & \#Dev & \#Test & \#Label \\\hline
CoLA   & 8.5K    & 1K    & 1K     & 2       \\
SST-2  & 67K     & 0.8K  & 1.8K   & 2       \\
MNLI   & 393K    & 20K   & 20K    & 3       \\
RTE    & 2.5K    & 0.2K  & 3K     & 2       \\
QQP    & 364K    & 40K   & 391K   & 2       \\
MRPC   & 3.7K    & 0.4K  & 1.7K   & 2       \\
QNLI   & 108K    & 5.7K  & 5.7K   & 2      \\\hline
\end{tabular}
\end{table}

\begin{table}[!htbp]
\caption{Summary of the Chinese benchmark}
\label{table:t2}
\begin{tabular}{llllc}
Corpus   & \#Train & \#Dev & \#Test & \#Label \\\hline
Chnsenti & 9.6K    & 1.2K  & 1.2K   & 2       \\
XNLI     & 392K    & 2.4K  & 5.0K   & 3       \\
ECDT     & 2.2K    & 0.6K  & 0.7K   & 31     \\\hline
\end{tabular}
\end{table}

GLUE, the General Language Understanding Evaluation, is a collection of six natural language understanding tasks that can be classified into two categories: single-sentence tasks and sentence pair tasks. We test our methods on seven of the nine tasks in the GLUE benchmark.

\noindent\textbf{Single-sentence tasks: }\\
In this category, we test  our proposed model on the following two tasks: Acceptability classification with CoLA \cite{Warstadt2018} and binary sentiment classification with SST \cite{socher2013recursive}.

\noindent\textbf{Sentence pair tasks:}\\
The following sentence pair tasks are used in our experiments: Semantic similarity with the MSR Paraphrase Corpus(MRPC) \cite{dolan2005automatically}, Quora Question Pairs (QQP) dataset, and textual entailment with Multi-Genre NLI Corpus MNLI \cite{williams2017broad}, a subset of the RTE challenge corpora \cite{dolan2005automatically}.

We exclude the Winograd NLI task and STS-B task. The experiments are conducted on the development set for some reason.

\subsubsection{Chinese Benchmark}
\noindent\textbf{ChnSenti} \newline
ChnSenti\cite{songbo} is a dataset which aims at judging the sentiment of a sentence. It includes comments in several domains such as hotels, books and electronic computers.

\noindent\textbf{XNLI}\\
The Cross-lingual Natural Language Inference (XNLI) corpus \cite{conneau2018xnli} is a crowd sourced collection for the MultiNLI corpus. The pairs are annotated with textual entailment and translated into 14 languages including Chinese. The labels contains contradiction, neutral and entailment.

\noindent\textbf{ECDT}\\
ECDT\cite{zhang2017first} is a dataset used in  the First Evaluation of Chinese Human-Computer Dialogue Technology. In using of human-computer dialogue based applications, human may have various intent, for example, chit-chatting, asking questions, booking air tickets, inquiring weather, etc. So the task is to classify the user intent in single utterance into a specific domain. The two top categories are chit-chat and task-oriented dialogue. Meanwhile, the task-oriented dialogue also includes 30 sub categories.

\subsection{Weight Initialization Strategy}
As we all know, the solution to a non-convex optimization algorithm  depends on the initial values of the parameters. In this part, we will discuss the weight initialization strategies of Transformer which is used as base classifier in BoostingBERT. We have four strategies: pre-trained initialization strategy, fine-tuning initialization strategy, incremental initialization strategy and random initialization strategy.

BERT \cite{devlin2018bert} is a new language representation model, which uses bidirectional transformers to pre-train a large corpus in the first stage and fine-tune the pre-trained model on specific task in the second stage.  The pre-trained initialization strategy uses the pre-trained model released in the first stage and fine-tuning initialization strategy initialize each base Transformer in BoostingBERT with the parameters after fine-tuning on the task on hand. Initializing weights of the number $n$ base Transformer with  the parameters of  number $(n-1)$ base classifier after it has been fine-tuned on the task, we call this approach \enquote{incremental initialization strategy}.  As a baseline to compare, the Xavier random initialization \cite{glorot2010understanding} is also adopted.

We will show the performance comparison of four initialization strategies in the experimental results section of this paper.

\subsection{The Impact of Pre-training Weight Initialization Strategy}

\begin{table}[!htbp]
\caption{GLUE tasks dev set results(different weight initialization strategies)}.
\label{table:t8}
\begin{tabular}{lllll}
Model & MRPC & MNLI-m & RTE & Score \\ \hline
Random & 70.34 & 65.83 & 54.15 & 63.43 \\
Pre-trained & 87.87 & \textbf{85.16} & \textbf{72.92} & \textbf{81.98} \\
Fine-tuning & \textbf{87.99} & 84.75 & 70.76 & 81.17 \\
Incremental & 86.52 & 84.63 & 70.76 & 80.64 \\ \hline
\end{tabular}
\end{table}

BERT shows excellent results on plenty of NLP tasks by leveraging large amount of unsupervised data for pre-training to get better contextual representations. In our proposed multi-class boosting  BERT model, we also investigate the impact of four different weight initialization strategies. Table \ref{table:t8} and Table \ref{table:t9} show the  performance comparison of different weight initialization strategies for some GLUE tasks and Chinese datasets respectively.
\begin{table}[!htbp]
\caption{Chinese dataset dev set results(different weight initialization strategies)}
\label{table:t9}
\begin{tabular}{lllll}
Model & ECDT & ChnSenti & XNLI & Score \\ \hline
Random & 69.18 & 86.75 & 58.67 & 71.53 \\
Pre-trained & \textbf{96.88} & 94.33 & \textbf{79.08} & \textbf{90.10} \\
Fine-tuning & 96.49 & 94.67 & 78.8 & 89.99 \\
Incremental & 95.71 & \textbf{94.92} & 77.83 & 89.49 \\ \hline
\end{tabular}
\end{table}

From the results, we can see that:
\begin{enumerate}
    \item Model initialized from pre-training weights outperforms training from scratch significantly. The relative performance improvement is around 29.24\% for GLUE and 25.96\% for Chinese benchmark.
    \item As for the comparison of  four weight initialization strategies, the English dataset and the Chinese dataset perform differently. For three typical GLUE tasks, the overall results show that the pre-trained weight initialization outperforms other weights initialization approaches. While there is no obvious winner  for Chinese datasets. The overall experimental results of pre-trained strategy and fine-tuning strategy show  similar performance, though  pre-trained approach is slightly better than fine-tuning weight initialization. The reason behind this still needs further exploration.
\end{enumerate}


\end{document}